\documentclass[letterpaper]{article}
\usepackage{aaai19}
\usepackage{times}
\usepackage{helvet}
\usepackage{courier}
\usepackage{graphicx}
\usepackage{xifthen}
\usepackage{amsmath,amsfonts,amssymb,amsthm}
\usepackage{color}
\usepackage{comment}
\usepackage[textsize=scriptsize,textwidth=1cm]{todonotes}

\usepackage{fancyvrb}
\usepackage{listings}

\usepackage{url}

\usepackage{tikz}
\usetikzlibrary{arrows, positioning, automata, shapes, calc}

\newcommand{\titlechoice}{Generating Dialogue Agents via Automated Planning}

\newcommand{\towrite}[1]{\hidecomments{\vspace{1em}\todo[inline,color=black!20!white]{\textbf{TODO:} #1}\vspace{1em}}}

\newcommand{\hidecomments}[1]{}

\newcommand{\ab}[1]{\hidecomments{\vspace{1em}\todo[inline,color=blue!20!white]{\textbf{Adi:} #1}\vspace{1em}}}

\newcommand{\cm}[1]{\hidecomments{\vspace{1em}\todo[inline,color=green!20!white]{\textbf{Christian:} #1}\vspace{1em}}}
\newcommand{\jo}[1]{\hidecomments{\vspace{1em}\todo[inline,color=yellow!20!white]{\textbf{Josef:} #1}\vspace{1em}}}

\newcommand{\ob}[1]{\hidecomments{\vspace{1em}\todo[inline,color=pink!20!white]{\textbf{Ondrej:} #1}\vspace{1em}}}

\newcommand{\optarg}[1]{\ifthenelse{\isempty{#1}}{}{(#1)}}

\newcommand{\hide}[1]{}

\newcommand{\la}{\langle}
\newcommand{\ra}{\rangle}

\newcommand{\fluents}{\mathcal{F}}
\newcommand{\actions}{\mathcal{A}}
\newcommand{\init}{\mathcal{I}}
\newcommand{\goal}{\mathcal{G}}

\newcommand{\nodes}{\mathcal{N}}
\newcommand{\edges}{\mathcal{E}}
\newcommand{\initnode}{n_0}

\newcommand{\actmap}[1]{act\optarg{#1}}
\newcommand{\goalaction}{{\mathbf{Done}}}

\newcommand{\problem}{\la \fluents, \init, \actions, \goal \ra}
\newcommand{\solution}{\la \nodes, \edges, \initnode \ra}

\newcommand{\pre}[1]{\mathrm{Pre}_{#1}}

\newcommand{\dialnodes}{\mathcal{V}}
\newcommand{\dialedges}{\mathcal{T}}
\newcommand{\dialinitnode}{v_0}
\newcommand{\dialgoals}{G}

\newcommand{\dialplan}{\la \dialnodes, \dialedges, \dialinitnode, \dialgoals \ra}

\theoremstyle{definition}

\newtheorem{definition}{Definition}

\def\intextcite#1{\citeauthor{#1}~\shortcite{#1}}

\title{\titlechoice}

\author{
Adi Botea \\ IBM Research \And
Christian Muise \\ IBM Research \And
Shubham Agarwal \\ IBM Research \And
Oznur Alkan \\ IBM Research \AND
Ondrej Bajgar \\ IBM Watson \And
Elizabeth Daly \\ IBM Research \And
Akihiro Kishimoto \\ IBM Research \And
Luis Lastras \\ IBM Research \AND
Radu Marinescu \\ IBM Research \And
Josef Ondrej \\ IBM Watson \And
Pablo Pedemonte \\ SilverGate -- IBM Argentina \And
Miroslav Vodol\'an \\ IBM Watson
} 

\hide{
\setlength\titlebox{3.5in}
\author{
Adi Botea \\ adibotea@ie.ibm.com \\ IBM Research \And
Christian Muise \\ christian.muise@ibm.com \\ IBM Research \And
Shubham Agarwal \\ shubham.agarwal@ibm.com \\ IBM Research \And
Oznur Alkan \\ oalkan2@ie.ibm.com  \\ IBM Research \AND
Ondrej Bajgar \\ ondrej@bajgar.org \\ IBM Watson \And
Elizabeth Daly \\ elizabeth.daly@ie.ibm.com \\ IBM Research \And
Akihiro Kishimoto \\ akihirok@ie.ibm.com \\ IBM Research \And
Luis Lastras \\ lastrasl@us.ibm.com \\ IBM Research \AND
Radu Marinescu \\ radu.marinescu@ie.ibm.com \\ IBM Research \And
Josef Ondrej \\ josef.ondrej@ibm.com \\ IBM Watson \And
Pablo Pedemonte \\ ppedemon@ar.ibm.com \\ SilverGate -- IBM Argentina \And
Miroslav Vodol\'an \\ mvodolan@cz.ibm.com \\ IBM Watson
} 
}

\nocopyright

\begin{document}

    \maketitle

\begin{abstract}
Dialogue systems have many applications such as customer support or question answering. Typically they have been limited to shallow single turn interactions. However more advanced applications such as career coaching or planning a trip require a much more complex multi-turn dialogue. Current limitations of conversational systems have made it difficult to support applications that require personalization, customization and context dependent interactions.
We tackle this challenging problem by using 
domain-independent AI planning to automatically create dialogue plans,
customized to guide a dialogue towards achieving
a given goal.
The input includes a library of atomic dialogue actions,
an initial state of the dialogue, and a goal.
Dialogue plans are plugged into a dialogue system capable
to orchestrate their execution.
Use cases demonstrate the 
viability of the approach.
Our work on dialogue planning has been integrated into a product,
and it is in the process of being deployed into another.
\end{abstract}

\section{Introduction}
\label{sec:intro}

Dialogue agents are becoming increasingly pervasive across many industries.
\towrite{$\langle \langle$ list of industry-specific agents $\rangle \rangle$.}
There is also a recognized demand for goal-oriented agents that are capable of bridging several services to assist human users in multi-turn dialogue \cite{DBLP:journals/aim/Ortiz18}. It is natural, then, to view the generation of dialogue agents through the lens of a technology well-suited for multi-step and goal-oriented settings: automated planning.

State-of-the-art dialogue systems 
typically can fall into two categories, dialogue trees and conversation learners. ELIZA is a first dialogue system that used dialogue trees~\cite{Weizenbaum:1966:ECP:365153.365168}.
In ELIZA and more recent tree-based dialogue systems,
the conversation is structured by a complex series of branches with options depending on the user responses, and the dialogue at each branching node is hard coded. Dialogue trees can rapidly become unwieldy as the choices become more complex. Additionally, a great deal of repetition may exist where similar dialogue interactions may be required at different points in the conversation.

Conversation learners use machine learning to compute the next response from examples or historical interactions \cite{DBLP:conf/ijcai/IlievskiMHB18,DBLP:conf/acl/GaoWPLL18}. They allow the
quick development of a complex dialogue without the laborious task of constructing a dialogue tree. Drawbacks include the lack of control from the application owner and the potentially unpredictable results. In certain domains, uncertainty and the risk of erroneous responses may be acceptable, but in domains such as health, finance and human resources (HR)
we require guarantees and predictability of results.

A modern dialogue agent should integrate important properties such as:
multi-turn, goal-oriented dialogues;
calls to external services whose output is needed in the dialogue;
and handling contingencies.
Informally, the latter covers the fact that, 
at certain points in a dialogue, the conversation
could steer in more than one direction.
For instance, the user could accept or reject
a suggestion from the agent,
and the follow-up dialogue
could be very different in each case.

%

We introduce an approach to constructing dialogue agents with AI planning.
Our dialogue plans feature the properties mentioned above.
These properties are not necessarily new,
but our novelty stems  from
computing such dialogue plans \textit{automatically},
using domain-independent planning.

This allows to create plans
tailored to specific scenarios
(e.g., focused on achieving a given goal).
The input includes a library of actions representing individual steps in a dialogue.
These are the building blocks used to construct dialogue plans.
As typical in AI planning,
the input further contains a problem instance,
which states the initial state and the goal of the dialogue plan to compute.
The system computes a dialogue plan that 
starts from the initial state and continues
until the goal is achieved.

A library of actions can be used to generate a range of dialogue plans,
and a given subset of actions can be used across several domains.
For instance, the user can ask about the weather in a range of domains,
such as trip planning and family event planning.
The ability to automatically compute dialogue plans
allows us to easily maintain deployed dialogue systems.
Fixing a bug, or slightly modifying the behaviour
of an action further implies the need to recreate
all dialogue plans impacted by the changes.
Manually fixing a collection of plans could be tedious and error prone.
Avoiding this is a key advantage to our approach.

Use cases show the feasability and the scalability of the approach.
In addition, our work has been integrated into a human resources product,
and it is being added to another product, for career coaching.

\section{Preliminaries}
\label{sec:preliminaries}




We introduce basic concepts needed in a dialogue system
capable to integrate our dialogue plans.
In the second part of this section we overview
fully observable non-deterministic planning (FOND).
This is the main planning approach we use,
together with elements of contingent planning.

\subsubsection{Dialogue} 
In a dialogue, user utterances are classified
into so-called \emph{intents}.
For example, a statement such as \emph{``I wonder how the weather is in Paris''}
could be classified into an intent such as {\sc \#asked-about-weather}.
We assume that intents are predefined (when the system is designed),
and the system is pretrained for intent classification.

\emph{Entities} are variables defined in a dialogue system.
Besides general-purpose variables,
such as places (e.g., \emph{Paris}), people names, dates and numbers,
designers can define entities specific to the domain at hand,
with a range of values that an entity can take.
A given value could possibly be defined in multiple ways, as a list of synonyms.
Value assignments to entities can automatically be recognized
in a user utterance, using 
Natural Language Understanding (NLU)~\cite{Manning:1999:FSN:311445,Jurafsky:2000:SLP:555733,Florian:2010:IMD:1870658.1870691}.
User utterances are annotated with recognized intents
and entity instantiations.
The sample utterance presented earlier
would be annotated with both the intent mentioned,
and with a variable assignment such as {\sc @place = Paris}.
As mentioned earlier, we allow calls to external services
in the middle of a dialogue.
Calls to external services are annotated with their inputs and outputs.

Such annotations will allow us to build more advanced \emph{context} information.
Informally, the context contains 
information that is relevant at a given point in a dialogue.
See details in Section~\ref{sec:architecture}.


\subsubsection{Fully Observable Non-deterministic Planning}

A \textit{FOND problem} is defined by a set of \textit{fluents} that
can be true or false in the environment, an \textit{initial state}
represented as a set of fluents, a set of \textit{non-deterministic
  actions} and a \textit{goal} condition that must hold at the end of
the plan. 

The FOND representation is similar to classical planning, except that
in FOND actions can have multiple effects (see also
\cite{DBLP:series/synthesis/2013Geffner} for additional details).

A \textit{non-deterministic action} consists of a
\textit{precondition} that must be satisfied for the action to be
applicable and a set of \textit{outcomes}, one of which will be used
during execution. Each outcome is a set of fluents or their negation,
indicating if the fluent should be added or deleted from the state of
the world.

Exactly one outcome 
occurs when an action is executed, and thus a solution to a FOND
problem must adequately handle every possible outcome.  Among several
equivalent ways to define a solution, we will use one resembles more
closely our dialogue plans.

Consider a new action $\goalaction$ whose precondition satisfies the
goal. A \emph{solution} to a FOND problem is a directed graph
$(\nodes, \edges)$ where $\nodes$ is a set of nodes and $\edges$ is a
set of directed edges. Each node $n \in \nodes$ is associated
with an action (including $\goalaction)$ and has one outgoing directed
edge for each outcome of its corresponding action. $\edges$ is the
union of all edges across all nodes. Nodes labeled by the
$\goalaction$ action have no outgoing edges. We call these leaf
nodes. A solution contains at least one leaf node and exactly one root
node without any incoming edges.

The action associated with the root node must be applicable in the
initial state. \textit{Reachable states} are recursively defined by
applying the action in a state reached at node $n$ (starting with the
initial state and root node, respectively). The outcome that occurs
during execution dictates both the following state of the world, and
the following node for the solution to proceed to. A solution must
additionally have the following properties: for every reachable state
and node pair $(s,n)$, the action corresponding to $n$ must be
applicable in $s$.  For every reachable node $n$, some leaf node $n'$
can be reached through some selection of outcomes.

In summary, a solution is a directed graph where the nodes correspond to actions the agent takes and edges correspond to how the uncertain environment responds. There must always be some path to arrive at a goal node.

\cm{Do we want to get into fairness at all? This would be the place...}


    \section{Dialogue Plans}
\label{sec:dial-plans}

\cm{I think this should be the target output that we shoot for. Keep the connection to planning slim, as that is the main bulk of the approach (i.e., how do we encode things to generate something that maps to our target dialogue plan. Phrase it as, ``a dialogue agent is represented by what we refer to as a dialogue plan...''}

\hide{
\begin{definition}
  A \emph{dialogue plan} is a structure $\dialplan$.
  $(\dialnodes, \dialedges)$ is a directed graph,
  with $\dialnodes$ being the nodes and $\dialedges$ being the edges.
  One unique node, labeled $\dialinitnode$, represents
  the \emph{initial node} of the dialogue.
  $G \neq \emptyset$ is the set of nodes with no outgoing edges,
  called the goal nodes.
  Nodes have two labels each: a \emph{condition label} and an \emph{action label}.
  A condition label is a Boolean formula where atoms are fluents from a set $\fluents$.
  An action label is a string.
\end{definition}
}

\begin{definition}
  A \emph{dialogue plan} is a structure $\dialplan$.
  $(\dialnodes, \dialedges)$ is a directed graph,
  with $\dialnodes$ being the nodes and $\dialedges$ being the edges.
  One unique node, labeled $\dialinitnode$, represents
  the \emph{initial node} of the dialogue.
  $G \neq \emptyset$ is the set of nodes with no outgoing edges,
  called the goal nodes.
  Each node has an \emph{action label} represented as a string.
  When a node has multiple outgoing edges, each such an
  edge has a Boolean formula where atoms are 
  fluents from a set $\fluents$.
\end{definition}

We use FOND planning
to generate dialogue plans.
Computing dialogue plans with AI planning
provides a mechanism
to construct the formulas associated to edges.
As the multiple edges originating from a node
are non-deterministic effects of a given action,
we use the outcome of each effect 
as the formula of the corresponding edge.

\begin{figure}
  \centering

  \definecolor{yellowfill}{RGB}{255,242,204}
  \definecolor{yellowborder}{RGB}{214,182,86}
  \definecolor{redfill}{RGB}{248,206,204}
  \definecolor{redborder}{RGB}{184,84,80}
  
  \tikzset{rectangular/.style={draw, rectangle, minimum height=1cm}}
  \tikzset{circular/.style={draw, circle}}
  \tikzset{yellow/.style={draw=yellowborder,fill=yellowfill}}
  \tikzset{red/.style={draw=redborder,fill=redfill}}

  \resizebox{.4\textwidth}{!}{

    \begin{tikzpicture}[
    shorten >= 1pt, 
    node distance=1.8cm, scale=0.5, semithick, ->]
  \tikzstyle{every node} = [font=\footnotesize]
  \node[
    state, 
    rectangular, 
    yellow, 
    align=center, 
    text width=2.7cm] (query-checkin) {\textsf{A: ask-checkin-luggage}};
  \node[
    state, 
    rectangular, 
    yellow, 
    below of=query-checkin, 
    align=center, 
    xshift=1.85cm, 
    text width=2.5cm] 
    (checkin-nonumber) {\textsf{A: ask-how-many}};
  \node[
    accepting, 
    rectangular,
    red,
    align=center, 
    text width=2.5cm, 
    below of=query-checkin, 
    xshift=-1.85cm, 
    double distance=1pt] (noalarm) {\textsf{A: Done}};
  \node[
    state, 
    rectangular,
    yellow, 
    align=center, 
    text width=3.8cm, 
    xshift=1.3cm, 
    below of=checkin-nonumber] (checkin-number) {\textsf{A: set-luggage-checkin}};

  \node[
      accepting, 
      rectangular,
      red, 
      align=center, 
      text width=2.5cm, 
      below of=checkin-number] (alarm-set) {\textsf{A: Done}};
  
  \draw[<-] (query-checkin) -- ++(-4cm,0);
  \path 
    (query-checkin) edge node[anchor=east, yshift=1mm]{\textsf{}} (checkin-nonumber)
    
    (query-checkin) edge[bend right=20] 
      node[anchor=east, xshift=1mm, yshift=2mm]{\textsf{}} (noalarm)
    
    (query-checkin) edge[bend left=60] 
      node[anchor=east, xshift=1mm, yshift=2mm]{\textsf{}} (checkin-number)

    (checkin-nonumber) edge[bend left=20] 
      node[anchor=west, xshift=-1mm, left, align=center]{\textsf{[have-number]}} (checkin-number)

    (checkin-nonumber) edge[loop above] node {[ ]} (checkin-nonumber)

    (query-checkin) edge[loop above] node {} (query-checkin)

    (checkin-number) edge[bend left=20] 
      node[anchor=west, xshift=-1mm, yshift=2mm]{\textsf{}} (alarm-set);

    \end{tikzpicture}
  }
  \caption{Toy dialogue plan. The plan has two goal states,
  shown as double-bordered boxes.
  Edge formulas shown for only one set of non-deterministic edges,
  to avoid clutter.}
  \label{fig-fsm}
  \end{figure}
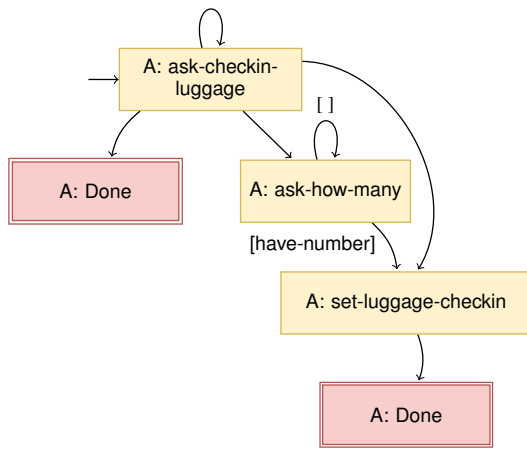

Figure~\ref{fig-fsm} shows a toy example of a dialogue plan
inspired from a trip planning application.
In this example, 
the agent asks the user whether the agent should 
check in any luggage in an upcoming flight.
The plan captures four possible options for the user's answer:
1) no luggage should be checked in, in which case the dialogue
can progress to the goal state at the left;
2) the user gives a positive reply, and provides the number of 
suitcases (e.g., \emph{``Yes, 2 pieces''});
3) the user gives a positive reply, without specifying the number;
and 4) the user gives an irrelevant response, in which case the agent
asks again.
In case 2, the dialogue can progress to calling an external service
that marks the corresponding field in a flight booking form,
after which the dialogue progresses to a goal state.
In case 3 the agent asks about the number. If a number is given,
the dialogue progresses to case 2.
Otherwise, the agent would ask again for a number.
For simplicity, we skip details such as interrupting the dialogue
after a finite number of iterations, in case that the user
keeps providing meaningless answers.

    \section{Architecture Overview}
\label{sec:architecture}

Here, we present an architecture that integrates
dialogue plans into an overall dialogue system,
starting with the key definition required for
maintaining the dialogue status.

\begin{definition}[Context]
    Given a set of variables $\cal W$,
    a \emph{context} is a partial instantiation of $\cal W$.
    In other words, a context contains instantiations to a subset 
    of the variables in $\cal W$.
\end{definition}

As mentioned, the context contains information available in the dialogue system at a given time.
Variables in $\cal W$ can include intents, entities, and variables to store
the outputs of calls to external services 
(i.e., variables instantiated as described in Section~\ref{sec:preliminaries}).
The context can contain additional variables, with rules about how and when
to instantiate them.
For instance,
a trip planning domain can define variables such as 
{\sc location-dest} and {\sc location-orig}.
During the dialogue, the context
can assign the automatically recognized value of the {\sc @place} entity
to either {\sc location-dest} or {\sc location-orig},
depending on the state of the dialogue.

Each goal that can be considered in some dialogue plan
has a corresponding intent in the dialogue system, called a \emph{top-level intent}.
User utterances classified into a top-level intent
trigger the execution of a dialogue plan with the corresponding goal.

Recall 
that a dialogue plan obtained from a planning system is
a directed graph.
A unique edge originating from a node represents
a deterministic transition, and multiple edges from a given node
represent the branches of a non-deterministic action.
%
Each action in the plan (represented as a string) needs to be mapped into an actual code to execute.
We call the code corresponding to an action a \emph{transformer}.
As such, each dialogue trace modelled in the dialogue plan
has a corresponding sequence of transformers to call.
As a dialogue progresses along a given trace,
the context can change after each step.
For instance, analysing a user utterance can lead to 
new entity instantiations.
Likewise, calling an external service leads to new output.
A transformer can 
consume context information (i.e., use context instantiations as input)
and produce context information (i.e., populate the context with new instantiations).
When the action at hand involves calling an external service
(e.g., call a career-pathway recommender system),
the corresponding transformer takes as an argument the 
link to the API of the external service.
The transformer calls the external service with the input at hand
(e.g., user profile stored as a context variable)
and places the results in dedicated context variables.

\towrite{Modify this accordingly, in case that the definition of a dialog plan changes.}
Consider a node $n$ in the plan, with multiple outgoing edges,
to a set of children nodes $c_1, \dots, c_l$.
When the dialogue continues from a node with 
multiple outgoing edges (branches), such as $n$,
the execution needs to decide what branch to choose.
That is, we need a mechanism to observe part of the dialogue state (context) and
make a decision based on that observation.
At the end of applying the action corresponding to node $n$,
the context allows to infer the current planning state.\footnote{To
    achieve this, each fluent from the planning problem is also defined as a context variable,
with a rule about how to instantiate (evaluate) it to {\sc true} or {\sc false}.
}
We use the current planning state, the formulas defined for
branches leading to the children nodes $c_1, \dots, c_l$,
and the previous planning state (when the execution was at node $n$),
to infer which branch should be followed.
We assume that the effects of exactly one branch are consistent
with the transition from the previous planning state to the current one.

Consider the example presented in Figure~\ref{fig-fsm}.
The node corresponding to the action {\sc ask-how-many} has two outgoing branches, 
corresponding to two non-deterministic effects of 
whether the user provides a number or not.
One branch is a self loop with no effects (i.e., no number provided)
and the other progresses to a different state, with the
number of luggage pieces given.

\hide{
Conditions defined in such a way are evaluated when the execution trace
reaches the parent state of the corresponding branches.
The evaluation is based on context information.
More specifically, each PDDL predicate used in a condition has associated an evaluation
formula based on variables from the context.
In our running example, the predicates {\sc user-accept} and
{\sc not-accepted} are evaluated based on 
the classification (intent) of the most recent utterance
(acceptance or rejection).
}

A deeper discussion on monitoring the execution of a plan is beyond the scope of this paper.

\section{A Planning Model for Dialogue}
\label{sec:approach}



\towrite{
  Mention somewhere the fact that 
  sometimes the dialogue has to be abandoned, if the user repeatedly
  misses the point (e.g., they repeatedly provide gibberish input, 
  or reject the recommendations for a predefined number of times).
  Describe a simple mechanism to implement this.
}

AI planning problems are often expressed in a domain-independent language from the PDDL family. 
Our PDDL models for different dialogue domains
share some commonalities in terms of the high-level design strategy.
We present lessons learned when designing dialogue planning models.

\subsubsection{Choosing the Level of Abstraction in PDDL}


The availability of the context,
separately from a dialogue plan,
eliminates the need to explicitly model in the PDDL problem description
all objects (i.e., possible values of variables) that could occur in a dialogue.
This abstraction avoids an artificial blow up in the problem size,
and in the solving effort.

Assume, for instance,
that at some point in a trip planning dialogue,
the destination has been set.
In the PDDL modeling, it is sufficient to encode that the destination
is known, with no need to explicitly name the destination.
That is, we use a predicate such as {\sc have-location-dest}, as opposed to 
{\sc have-location-dest ?loc}.
The latter would have to be instantiated into many grounded fluents,
one for each possible destination.
In contrast, the former is grounded into exactly one instantiated fluent,
with corresponding savings in the problem size and difficulty.

A PDDL model abstracts away some, but not necessarily all information included in the context.
As mentioned in Section~\ref{sec:architecture},
part of the predicates used in the PDDL model 
are mirrored with corresponding variables in the context,
to be able to decide on what branches to continue with the 
execution of a dialogue plan.

\subsubsection{Basic Fluents} Following the previous discussion on using the right level
of abstraction, we introduce the following types of fluents for the PDDL model:

\begin{tabular}{rl}
{\sc ok-*} & To indicate if a Boolean flag holds true. \\
{\sc have-*} & To indicate we have a context value. \\
{\sc maybe-*} & To indicate uncertainty of a context value. \\
{\sc goal} & Specially designated fluent for the goal.
\end{tabular}

\cm{I reduced {\sc maybe-have-*} to {\sc maybe-*} for space reasons, but perhaps we could
use {\sc know-*} and {\sc poss-*} instead of {\sc have-*} and {\sc maybe-*} (respectively)
throughout the paper.}

For a context variable, such as {\sc location-dest}, the fluents {\sc have-location-dest}
and {\sc maybe-location-dest} make a 3-valued logic (at most one can be true). We found
the latter mode to be essential for tailored dialogue that responds appropriately to
uncertain data (e.g., asking ``\textit{You'll be traveling to Berlin, right?}'' instead
of ``\textit{Where will you be traveling to?}'' when {\sc maybe-location-dest} holds and
we have some idea what the location should be).

The {\sc goal} fluent captures the fact that we typically achieve the dialogue goal by
means of executing a particular action (e.g., booking a trip or making a successful
recommendation). We elaborate on this further in Section \ref{sec:multi-intent}.

\subsubsection{Basic Actions}
We have identified two key action types that
are shared across the dialogue domains: dialogue actions and service actions.
The model could optionally include other actions, such as an auxiliary action at
the end of every plan, to indicate the termination of the dialogue, but the two
pervasive types are what we discuss.


\textit{Dialogue actions} correspond to sending messages to the end-user
in a conversation.
We assume that the executor of a plan has a way to map a given
dialogue action (along with the current context) to an utterance that should be
sent to the end user. 
If the dialogue action has more than one outcome, it is
presumed to be a message that warrants a response from the user, and the user's
response will correspond to the various action outcomes.

\ab{The next paragraph might belong to a different section.}

In the deployed dialogue plans we have created, the execution of a dialogue action
sends the message to the end-user using a common messaging protocol, and the response
in situations with more than one outcome is assessed using off-the-shelf NLU technology
(e.g., services for natural language disambiguation and entity extraction). The effects
of an outcome for a dialogue action can encode \emph{whether} various types of information are
available.
Separately from the dialogue plan,
the context will store the actual values of those types of information.

Eliciting information from the user is an important feature in multi-turn, goal-oriented
dialogues. Dedicated fluents encode whether a given type of information has successfully
been elicited. Fluents such as {\sc have-employee-name} (in a HR 
dialogue where a manager can ask about the performance of various team members)
and {\sc have-location-dest}, mentioned
earlier, are prime examples of this.
Once again, during execution, the context is updated to reflect
the actual values that have been acquired or modified.

The preconditions of a dialogue action dictate \textit{when} such an utterance
or question would be posed to the user. For example, querying a user's destination location
only makes sense if we do not already have it. More subtly, an action such
as {\sc ask-user-dest} would be predicated on \textit{not} having {\sc maybe-location-dest}
hold in the state, as we would instead prefer the action {\sc confirm-user-dest}.


\textit{Service actions} refer to the actions in the model that do not directly correspond
to messages that are sent to the end-user. These include system checks that have multiple
outcomes associated (essentially embedding key components of logic into the process of
compiling the dialogue agent) or even web API calls that may be required as part of the
conversation. An example would be looking up the weather using an online RESTful service.
The outcomes of a service action correspond to the possible responses we might expect and
wish to handle as part of the conversation.

The specific implementation details are beyond the scope of this paper, but essential to
a service action being used as part of a plan, we assume that the executor is capable of
making the RESTful API calls (or similar such service actions), and resolving the outcome.
Part of this resolution process is to update the context with new information as appropriate,
and maintain the corresponding state of the world from the view of the planner's abstraction.


\ab{The following two paragraaphs might belong 
to the section on abstraction levels.}

It is worth emphasizing the role of the outcomes from the dialogue designer's perspective.
There may be countless ways that a user could respond to a question, and similarly countless
error codes that a RESTful endpoint might return. However, the task of the dialogue designer
is to \textit{only specify the outcomes that contribute to changes in state and/or conversation}.
This means that a large variety of possible outcomes are categorized together.

An example for the service action might be mapping all error codes of the weather service
into one {\sc no-weather-service} outcome.
An example for a dialogue question might
be all of the ways the user could respond in the affirmative. It was a prevailing design
philosophy of the dialogue agent modelling that we should only consider the outcomes that
are \textit{required for conversation}, and including a catch-all outcome as needed when
the response is unclear (e.g., when the NLU cannot understand the end-user response).

\section{Advanced PDDL Features}
\label{sec:extensions}

\ab{I suggest to merge this with the previous section.
It's hard to draw a line between what is advanced and what is not.
Or, if the section would be too long, restructure on another
criterion than advanced/basic.}

\cm{The split is done assuming the standard form of a theoretical correspondence
between two systems. You don't want to show the entire model (including the syntactic
sugar that this section represents) and have to draw the correspondence with it all.
Instead, you present the core ideas, establish the connection, and then show the
extensions while claiming that it doesn't change the correspondence in any way
(i.e., the syntactic sugar argument). If we forgo any rigor on the model itself,
then I think it makes perfect sense to combine (as long as it still can read well
without reader fatigue).}

Having the base encoding in hand, we now describe some of the advanced encoding
features that we have identified and deployed for the dialogue agents we have created.
These stem from common patterns observed in addressing the pain points of dialogue
designers.

\subsubsection{Forced Followup}

Generally speaking, the declarative nature of planning can offer massive savings
to the process of dialogue design (and we demonstrate as such later in Section \ref{sec:eval}).
That said, there are some limited forms of imperative-style specification that
we found to be common in the domains we have investigated. Almost exclusively,
these took the form of immediate followup functionality: examples include responding
quickly with an affirmation, running a complex service action with many outcomes
after a particular response was received, etc. Here, we detail the modelling
strategy used for such situations.

\textit{Forced followup} is a modelling feature that introduces a new set of
fluents that are incorporated into the actions in a particular way. We use two
new fluents: (1) {\sc forced-followup-$t$} indicates if there is a forced followup
that must occur of type $t$; and (2) {\sc force-reason-$r$} indicates what the
reason is for the forced followup.

Examples we have considered for type $t$ include {\sc dialogue} (immediately
respond with a message), {\sc check} (run a system check), and {\sc abort} (to
abort the conversation and hand off to a human operator). The type of forced
followup allows us to predicate some subset of the actions with the ability
to handle the forced followup. If only one action can handle a particular type,
then the model should ensure that it is the \textit{only} applicable action.
We achieve this by having the negation of all types that an action $a$ cannot handle
as a precondition for $a$. For many of the actions, there will be no
type of forced followup that they can handle, which means they have a negated
precondition for every type (easily specified using quantified preconditions in PDDL).

We assume that actions which handle a particular type of forced followup
always remove the appropriate fluent as part of their effects (i.e., the
{\sc forced-followup-$t$} is deleted in every outcome of the action). This
ensures that the remaining actions in the domain are subsequently re-enabled.

The {\sc force-reason-$r$} fluents in some sense mirror the {\sc forced-followup-$t$}
fluents, as they are both added and deleted at the same time, but they additionally provide a higher
fidelity to the followup mechanism. As a grounded example, one of the domains
(discussed later in 
Section~\ref{sec:eval})
uses {\sc dialogue}
as a type for forced followup with reasons spanning a range of errors (such as
{\sc bad-weather}, {\sc bad-dates}, etc), warnings (e.g., {\sc no-weather-service}),
and affirmations (e.g., {\sc affirm-ok}). The action description in the model
makes use of the lifted representation for PDDL, and thus only one action is needed
to handle the range of forced responses corresponding to each reason. The
example action schema for a forced followup of type $dialogue$ would be
{\sc (handle-forced-dialogue ?r - reason)}

We found that the task of declaratively specifying a dialogue agent was greatly
simplified by allowing for this single-step imperative pattern to be used directly.
It essentially empowers the dialogue designer to specify the immediate followup
for key outcomes on certain actions, and additionally had the benefit of simplifying
the execution of the dialogue plans (as utterances from the agent to the end-user
need not be placed on outcomes).

\subsubsection{Handling Multiple Intents}
\label{sec:multi-intent}

If desired, one can build a dialogue plan with a 
disjunctive goal. Such a plan would satisfy any
one of a collection of goals (top-level intents)
considered in the disjunctive goal.
In such a case,
we introduce additional auxiliary fluents and actions to
address them: for each intent $i$, we have a fluent {\sc intent-$i$} that indicates if
the user has that intent, and a corresponding action {\sc assert-intent-$i$} is introduced
with the following properties:

\begin{enumerate}
  \item Only applicable when both {\sc intent-$i$} and the necessary condition for
        intent $i$ to be satisfied holds.
  \item Adds the goal fluent {\sc goal} as its only effect.
\end{enumerate}

In the domains we have experienced, it is enough to assume that only
a single intent needs to be confirmed.

\hide{
A common pattern in creating multi-turn dialogue agents is to model multiple possible
\textit{intents} that the user might have.\footnote{An important distinction
is that we refer to the top-level intents of a user, and not the common use in dialogue
literature i.e., the classification of every end-user utterance as an intent.}
When multiple intents exist, we introduce additional auxiliary fluents and actions to
address them: for each intent $i$, we have a fluent {\sc intent-$i$} that indicates if
the user has that intent, and a corresponding action {\sc assert-intent-$i$} is introduced
with the following properties:

\begin{enumerate}
  \item Only applicable when both {\sc intent-$i$} and the necessary condition for
        intent $i$ to be satisfied holds.
  \item Adds the goal fluent {\sc goal} as its only effect.
\end{enumerate}

The astute reader will recognize the disjunctive nature of this encoding method: if
the user has more than one intent that is recognized, satisfying only one of them would
suffice. However, in the domains we have experienced, it is enough to assume that only
a single intent need be confirmed. If multiple intents were of interest simultaneously,
then a new intent that represents their combination could be used.
}

    \section{Use Cases}
\label{sec:eval}

We evaluate our generic approach to computing dialogue plans with AI planning
in four domains:
human resources, career coaching, trip planning,
and a synthetic domain.
The first two correspond to two products.
The synthetic domain allows to 
perform a more focused scalability evaluation.

\ab{Differences in size/effort compared to more manual approaches. Adapting/maintainig/extending the system with ease. Any user study available? That would help too.}

\cm{Another is to measure the shift a small change in model can have to the final solution. Minor things in the declarative setting can lead to \textit{massive} changes in the contingent plan (which would be manual edits the way dialogue agents are currently designed).}

\cm{There is already a todo item to fix up the figure for the first two use cases --
would it be possible to switch the order of action / state? So nodes correspond to actions
while the state is implicit (then the outgoing edges correspond to outcomes of the action.}


\towrite{
    Introduce use case: relevant intents, building blocks, how building blocks can 
    pop up in multiple intents. Present the limitations of Watson Assistant,
    which leads to the need for specialized apps, such as Java apps. Then present the
    limitations of apps with hardcoded plans. More work to develop, more work 
    to maintain. Which leads to our approach, with one app that performs
    planning, planning monitoring and planning execution. Give some stats about
    the size of the apps.
}

\subsubsection{Human Resources} We present a HR application with
dialogues related to employee professional performance.
Regular employees could have a dialogue about their own performance.
In addition, a manager could have a chat about 
the performance of any team member,
and the team as a whole.

The performance is defined along a number of criteria called capabilities, such as
technical skills and knowledge.

In this domain, dialogue snippets can be partitioned into three main categories:
questions with static answers related to the domain
(e.g., \emph{``How can I use this tool?")};
general-purpose chitchat
(e.g., \emph{``Hello"});
and goal-oriented, multi-turn conversations.
The first two categories are simple snippets of one question and one answer.
The third one is handled with the approach presented in this paper.

Goals considered in multi-turn dialogues include:
(1) presenting the performance rating of a given individual for a given capability;
(2) giving an explaination about the value of a performance rating for an individual and a capability;
(3) recommending learning resources to improve the performance of a given individual in a given capability;
(4) presenting the performance rating of a team for a given capability;
(5) giving an explaination about the value of a performance rating for a team and a capability;
and
(6) identifying the strongest and the weakest performer in a team, for a given capability.

Such goals often require a multi-turn conversation,
for person disambiguation and capability disambiguation.
For instance, a person name given in the user utterance
is used to identify a person record in the team member database.
A first name provided in a user utterance might correspond to zero, one or several person records in the database. In the first and the third case, the dialogue needs to continue with disambiguating the
person.
Likewise, when no capability is specified, or the capability is ambiguous, the dialogue needs to 
disambiguate it.

An early implementation of the goal-oriented conversations used one
Java program for goals 1 and 4; one Java program for goals 2 and 5;
one Java program for goal 3; and one Java program for goal 6 (using a
total of 3,702 lines of code).  These Java programs are essentially
implementing hard-coded dialogue plans.  This is hard to maintain and
to extend to new goals, and the portability to a different domain is
very limited.

We have replaced such multiple individual Java apps, based on
hardcoded specific dialogue plans, with one single application (using
only 1,166 lines of code).  The application analyzes the goal at hand,
identified from the intent of the user utterance,
and constructs a planning instance accordingly.
The instance can be fed into an off-the-shelf planning system to obtain a dialogue plan on demand.
Alternatively, dialogue plans can be precomputed and stored into a library,
indexed on the goals they address.
The dialogue plan is executed and monitored in the system.



\subsubsection{Career Coaching} We consider two dialogue goals in career coaching:
reaching a point where the user has eventually chosen a long-term career goal;
and eventually chosing a career pathway towards that goal.
Thus, the system implements calls to two APIs: one for a career goal recommender, and one for a pathway recommender.
Each API implements calls that provide: a list of recommendations; an explanation associated with a given item (goal or pathway) included in a recommendation; additional details about a given item included in a recommendation.

For long-term career goals,
a service call to a recommender system provides a number of recommendations.
These are ordered, and the top three are presented to the user.
The user can choose a career goal,
or request additional information
(e.g., an explanation of why a given career goal is included),
or reject the career goals currently provided.
In the first case, the goal is achieved.
In the second case, additional information is provided and the dialogue continues
recursively, from the state where the user is required to choose between the three options again.
In the third case, additional information is elicited from the user, regarding the reason
of not liking any career goal. Based on the newly elicited information, the short list of
three recommendations is re-computed, and the dialogue continues recursively.

\begin{figure}
    \centering
    \includegraphics[width=.27\textwidth]{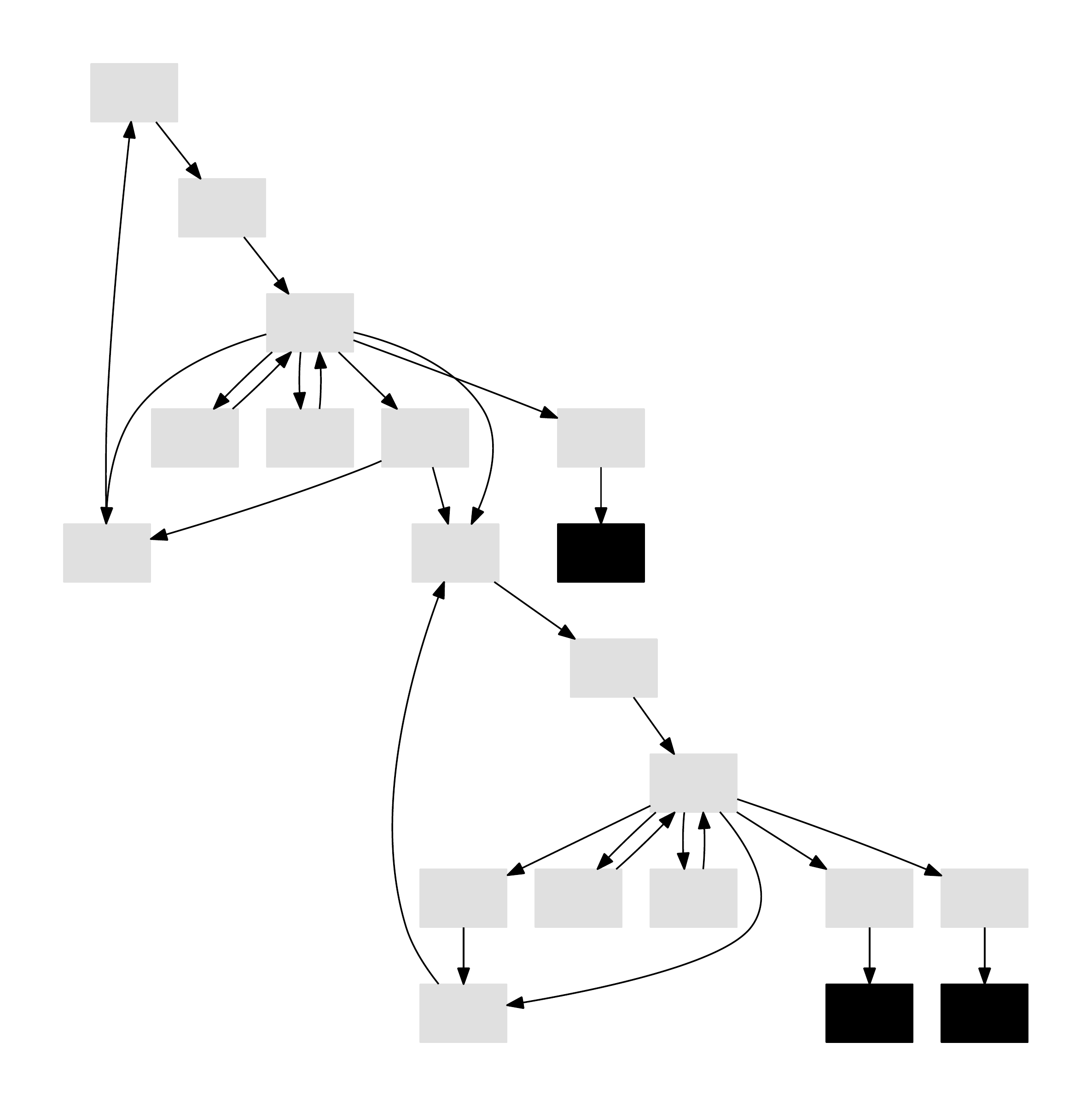}
    \caption{Dialogue plan for career goals and pathways.}
    \label{fig::myca-plan}
\end{figure}

\hide{
\begin{figure}
    \centering
    \includegraphics[width=.3\textwidth]{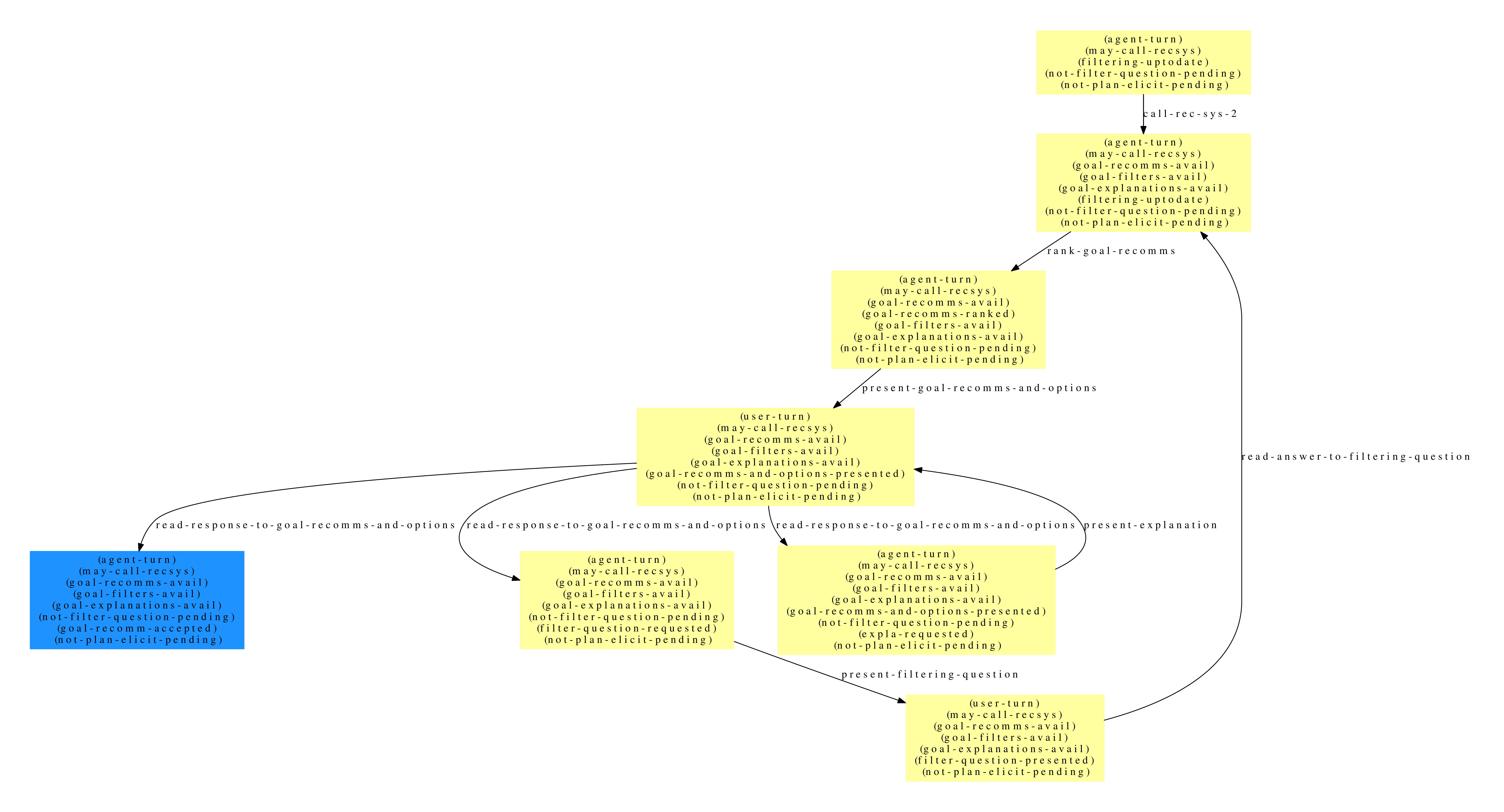}
    \caption{Dialogue plan focused on successfully recommending a long-term career goal to a user.}
    \label{fig::myca-cg-plan}
\end{figure}
}

Another dialogue plan focuses on helping the end user choose a career pathway.
Career pathways are computed with a call to an external service.
Pathways are presented to the user, which can accept or reject the career pathway at hand.
In the former case, the goal is reached and the dialogue concludes.
In the latter case, information about the reason of rejecting the pathway is elicited from the user.
Potential reasons can be related to a job role along the pathway, or constraints
associated with roles (e.g., the user might dislike management roles).
The context is updated, a career pathway is recomputed, and the dialogue
continues recursively.

\hide{
In practice, dialogue systems should implement the option to conclude the conversation
if no progress can be made after a given number of repetitions of a cycle in the flow.
An example is the case when the user keeps rejecting the recommendations.
Such an exit mechanism can be implemented, for instance, at the level of the execution monitoring,
by counting the executions of a cycle.
}

Computing a career pathways towards a long-term career goal requires the career goal as an input.
As such, the two dialogues could be chained in a sequence. They can also be generated
as independent dialogue plans. Such variations can easily be obtained with very small
modifications in the PDDL problem instance definition.
For instance, if the goal is to choose a career pathway, and a career goal is already
available in the initial state, there is no need to 
have the dialogue focused on choosing a career goal.
Otherwise, the two plans will automatically be chained in one larger dialogue plan.
Figure~\ref{fig::myca-plan} illustrates this combined dialogue plan.

\hide{
\begin{figure}
    \centering
    \includegraphics[width=.3\textwidth]{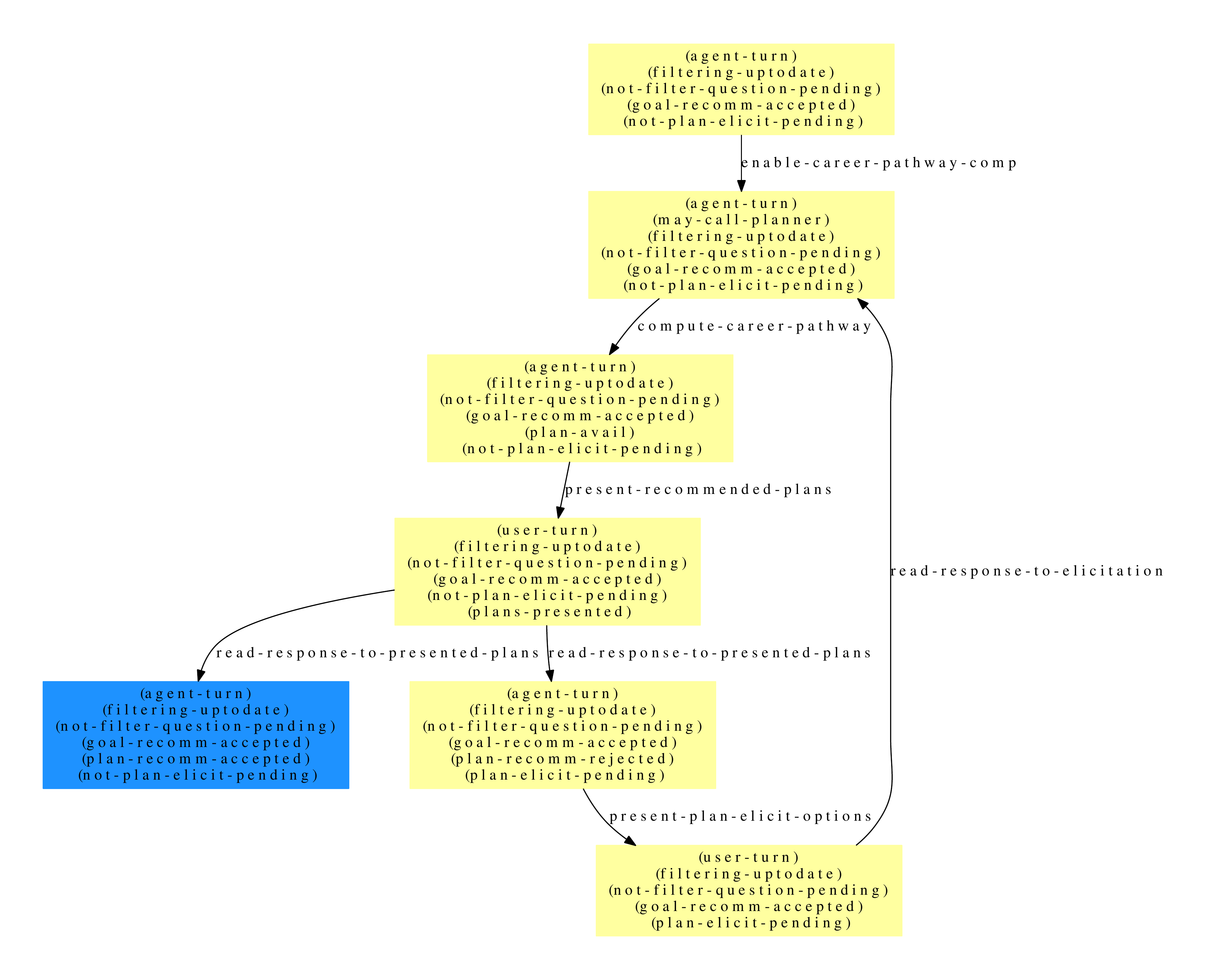}
	\caption{Dialogue plan focused on successfully recommending a career pathway to a user.
	Dialogue actions and service actions are visualised with different graphical symbols.}
    \label{fig::myca-cp-plan}
\end{figure}
}

\ab{TODO: Add an argument about the value of the work in this domain.}


\begin{figure}
    \centering
    \includegraphics[width=.4\textwidth]{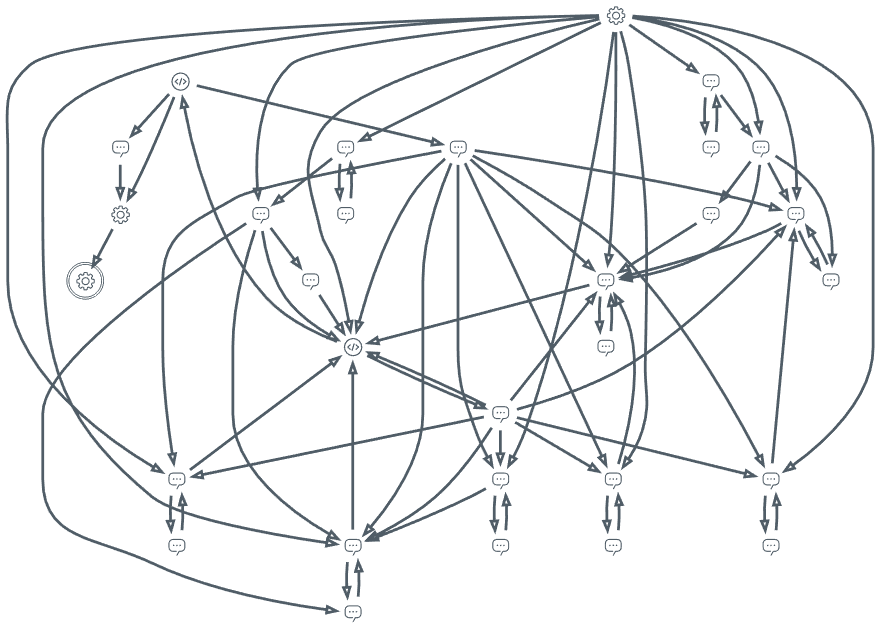}
    \caption{Generated dialogue for the Trip Planning use case.}
    \label{fig:trip-plan-example}
\end{figure}

\subsubsection{Trip Planning} The goal of our trip planning system is
to provide booking services while considering weather situation
relevant for the trip.  The system needs to collect departure and
arrival destinations and date range of the trip. We provide in the
supplementary material the actual PDDL domain and problem instance
used to drive the dialogue. For the previous two use cases, the PDDL
specification is closed.

The system is requesting the trip parameters through the natural dialogue and verifies their correctness (e.g. validity of provided locations).
After enough information is collected, the system uses an external call to check the weather situation for the specified destination and dates.
If the weather is evaluated as inferior, the system informs the user and suggests changing the trip parameters.

From the automated planning perspective, the interesting aspect of this use case is handling the uncertainty of collected system parameters.
The uncertainty comes from two sources: (1) the location recognition can be ambiguous due to NLU errors; and (2)
the system can hypothesize about arrival and destination locations based on historical data and the actual user location.
This leads to the planner to confirming information with the user that it has some certainty about, and soliciting information
from scratch when it does not have a sense as to what the true value is.

Figure \ref{fig:trip-plan-example} shows a high-level view of the generated dialogue plan that comes from a model with only
nine actions. The node symbols indicates the type of action that corresponds to that part of the plan: either dialogue, API
call, or system action (the latter two being specific examples of non-dialogue actions discussed in 
Section~\ref{sec:approach}).
Even in this limited setting, we can observe how complex behaviour can be captured in the generated dialogue agent from a simple
declarative specification.

\subsubsection{Scalability Analysis}

\hide{
\ab{Looks nice. A couple of comments: 
\begin{itemize}
	\item Is it possible, by any chance, to measure the solution differences in terms of the Graph Edit Distance (GED)?
	\url{https://en.wikipedia.org/wiki/Graph_edit_distance}
	\item Maybe the reviewers wonder why we have a sudden jump from 20 to 40 in Figure~\ref{fig:SRCbyIGC}. I.e., dot isolated at the bottom-left corner. Any insights?
\end{itemize}
}

\cm{\begin{itemize}
    \item I don't think so. We'd need to build a search algorithm from scratch to go from one graph to another. I cane envision how to roll this out from scratch (but don't have the time), and can't see a clear path to modeling it as planning (so we could just use a planner as the A* search). It's a very good measure (nice find!), but don't think it's feasible at the moment. Do you think we should cite it?
    \item Good point. It was from a single outlier that has only a small change in the solution for a fairly different init/goal. 7 things changed in the init/goal and only 134 lines changed in the solution. I've added text to this effect.
\end{itemize}}
} 


Scalability 
is a major advantage
of using a declarative representation for goal-oriented dialogue agents. To demonstrate this empirically, we created synthetic domains and problems mirroring the properties we observed in the existing dialogue encodings of our three use cases above. We measure the model size of the generated problems and solution size of the computed dialogue agents as the number of unique actions used in the solution and the total number of actions in the dialogue plan respectively.

We setup our experiments by populating a domain with random non-deterministic actions and a problem with random initial and goal states. Action precondition and effects are generated through random sampling from a set of fluents. We mirror the characteristics of actions inherent to a real dialogue system by: (1) keeping size of action preconditions within the range of 1-5 inclusive; (2) randomly sampling the effect type as either \textit{select} (exactly one of 2-5 fluents will become true) or \textit{assign} (1-4 fluents are randomly flipped); (3) randomly sampling 1-5 fluents for the initial state; and (4) randomly sampling 1-2 fluents for the goal state. As parameters to the random problem generator, we provide the number of actions and fluents.

The two action types correspond to typical dialogue actions that we have encountered. The \textit{select} action mirrors the determination of one type of response the user could provide (typically with a range of 2 to 5 possibilities). The \textit{assign} action mirrors dialogue actions where many aspects of the context can be assessed simultaneously (and thus multiple non-deterministic aspects are considered simultaneously). We have additionally confirmed qualitatively that the generated plans appear to contain a similar structure as the known dialogue plans (e.g., with the same expected plan length).

\begin{figure}
	\centering
	\includegraphics[width=.36\textwidth]{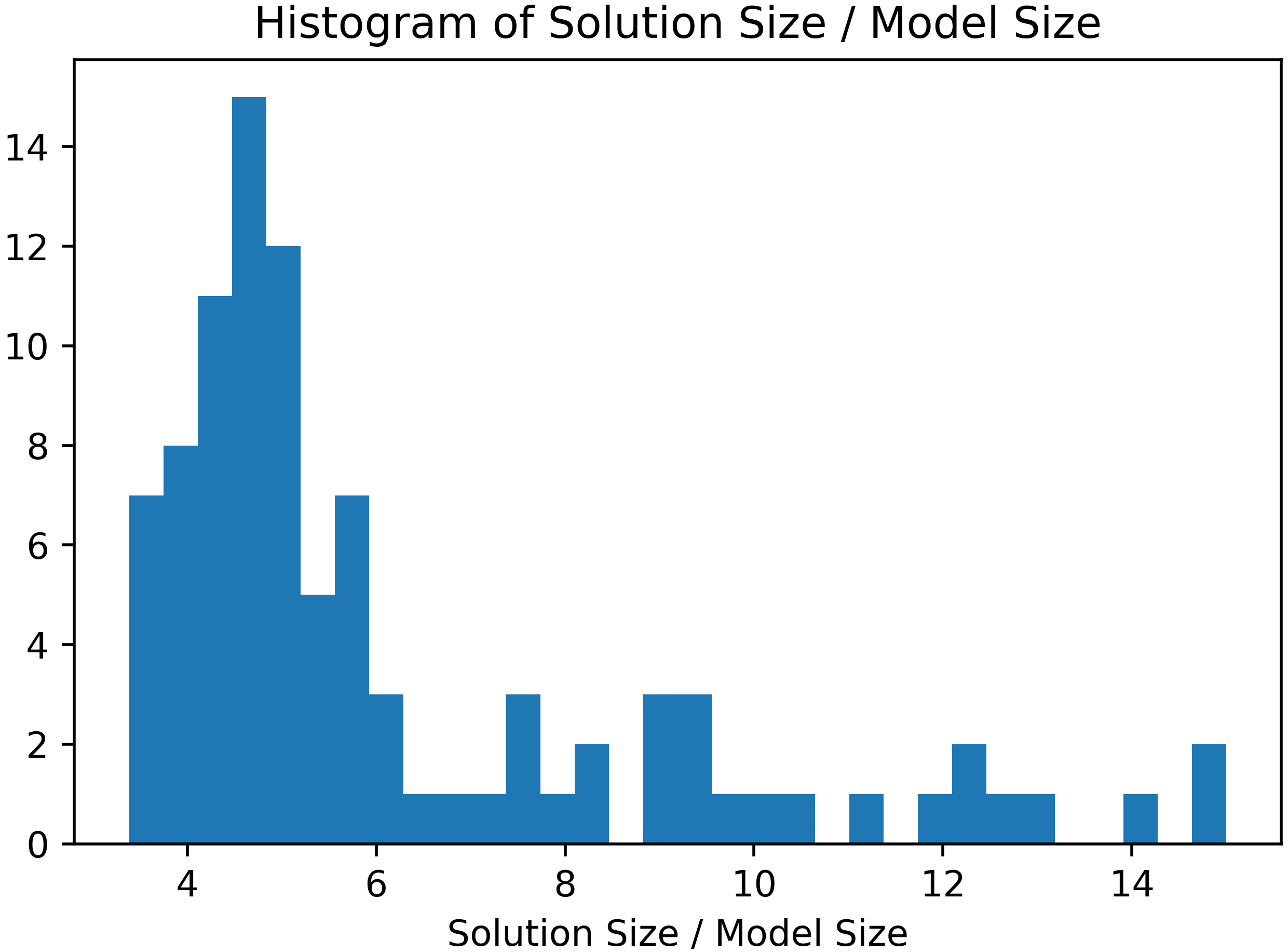}
	\caption{Plot of ratio of Solution size and Model size}
	\vspace{-1em}
	\label{fig::SSbyMS}
\end{figure}

In total we generated 100 instances. 
%
Figure \ref{fig::SSbyMS} shows a histogram of the ratio of solution size divided by model size.
In most of the instances, solution size is at least 4 times the size of the model, and in extreme cases it can grow to 16 times the model size. This confirms our assertion that complex dialogue systems can be efficiently designed with very compact declarative representations.


\section{Related Work}
\label{sec:related}



In the past, \citeauthor{Kuijpers1998}~\shortcite{Kuijpers1998}
presented an approach where a pre-existing library of plans can be
used by an agent in a
dialogue. \citeauthor{Steedman07planningdialog}~\shortcite{Steedman07planningdialog}
advocated the use of AI planning to facilitate mixed-initiative
collaborative discourse. More recently,
\cite{DBLP:conf/sigdial/NothdurftBBBM15} described a system capable of
explaining the decisions of the planning system to the
user. \intextcite{DBLP:journals/aim/Ortiz18} presented a holistic
approach to building conversational assistants which includes a
planning component to assist the interaction with human
users. However, none of these previous approaches employ AI planning
to generate customized dialog plans.

\citeauthor{garoufi2010automated}~\shortcite{garoufi2010automated}
showed recently how to generate whole sentences word-by-word with AI
planning techniques and described an approach to generate multi-turn
navigational commands to be followed by a human user in a
conversational interface to achieve a specified goal.

\intextcite{black2014automated} developed a formalism that describes
persuasion dialogues as state transitions in a planning domain.  In
this setup, an automated planner was able to find optimal persuasion
strategies. However, the execution monitoring is greatly simplified
due to the abstract nature of simulated dialogues.

\intextcite{petrick2013planning} considered execution monitoring
issues that arise from non-determinism of real-world dialogue systems.
Their proposed execution monitor expects deterministic behavior of
actions in the world and rebuilds the plan if inconsistencies between
measurements and the expectations are observed (via various sensors).
In contrast, our execution monitor accounts for non-determinism
explicitly within the generated plan. Therefore, our plans can be
static and easier to debug.

Another direction of research is focused on using (deep) reinforcement
learning
for end-to-end dialogue
generation \cite{bhuwan2017,peng2018}. In contrast to our model-based
approach, training these data centric systems is costly and
requires many interactions with human users.

The work that perhaps is the closest in spirit with ours is that by
\cite{Williams2007,Thomson2007,bui2010} who focus on using (factored)
POMDPs to manage spoken dialogue systems in various domains. However,
these systems, which are quite sensitive to the uncertainty inherent in
the spoken utterances, require considerable training to learn
good policies and thus drive the dialogue.



\section{Summary}
\label{sec:summary}

Dialogue agents capable of handling multi-turn, goal-oriented conversations
are becoming increasingly important across a range of domains, 
including human resources, career coaching, and personal assistants.
We have presented an approach to constructing dialogue plans automatically.
Given a library of individual actions available, 
our system produces dialogue plans customized to achieving a given goal.
Dialogue plans are further plugged into a dialogue system that can 
orchestrate their execution during a conversation with a user.
We have shown that our approach is viable and scalable.
Our work has been applied in one product, and is in the process of 
being integrated into a second one.

Future work includes building dialogue agents in additional domains.
Dynamically interleaving dialogue agents, to allow the user
to temporarily change the topic in the middle of a dialogue,
is another important topic for future work.


    \bibliography{references}
    \bibliographystyle{aaai}

\end{document}